\documentclass[9pt,conference]{IEEEtran}
\IEEEoverridecommandlockouts

\usepackage{cite}
\usepackage{amsmath,amssymb,amsfonts}
\usepackage{algorithmic}
\usepackage{graphicx}
\usepackage{textcomp}
\usepackage{xcolor}
\usepackage{comment}

\usepackage{array}
\usepackage{mathtools}
\usepackage{tcolorbox}
\usepackage{color}
\usepackage{theorem}
\usepackage{amssymb}
\usepackage{caption}
\usepackage{subcaption}
\usepackage{cite,hyperref}
\usepackage{cases}
\usepackage{url}
\usepackage[algoruled,linesnumbered,onelanguage]{algorithm2e}
\usepackage{algorithmic}
\usepackage{tikz}
\usetikzlibrary{shapes,arrows}
\usepackage{enumitem}

\usepackage{stmaryrd}

\usepackage{multirow}
\usepackage{balance}

\usepackage{tikz}
\usepackage{moresize}
\usepackage{pgfplots}
\usepackage{wrapfig}
\input{mysymbol.sty}
\pgfplotsset{compat=1.17}
\pgfplotstableset{col sep=comma}
\begin{document}

\title{A Tensor Low-Rank Approximation for Value Functions in Multi-Task Reinforcement Learning
\thanks{Work partially supported by the Spanish AEI (AEI/10.13039 /501100011033), grants TED2021-130347B-I00 and PID2022-136887NB-I00, the Community of Madrid via the Ellis Madrid Unit and the CAM-URJC F840 grant. and by the Office of Naval Research
(ONR), under contract number N00014-23-1-2377.
Emails:  \href{mailto:s.rozada.2019@alumnos.urjc.es}{s.rozada.2019@alumnos.urjc.es}, \href{mailto:paters@rpi.edu}{paters@rpi.edu}, \href{mailto:juanbazerque@pitt.edu}{juanbazerque@pitt.edu}, \href{mailto:antonio.garcia.marques@urjc.es}{antonio.garcia.marques@urjc.es}
}
}

\author{
\IEEEauthorblockN{
Sergio Rozada\IEEEauthorrefmark{1},
Santiago Paternain\IEEEauthorrefmark{2},
Juan Andrés Bazerque\IEEEauthorrefmark{3},
Antonio G. Marques\IEEEauthorrefmark{1}
} %
\IEEEauthorblockA{
\IEEEauthorrefmark{1}Dept. of Signal Theory and Communications, King Juan Carlos University, Madrid, Spain } %
\IEEEauthorblockA{
\IEEEauthorrefmark{2}Dept. of Electrical, Computer, and Systems Engineering, Rensselaer Polytechnic Institute, Troy, NY, USA } %
\IEEEauthorblockA{
\IEEEauthorrefmark{3}Dept. of Electrical
and Computer Engineering, University of Pittsburgh, Pittsburgh, PA, USA } %
}

\maketitle

\begin{abstract}
    In pursuit of reinforcement learning systems that could train in physical environments, we investigate multi-task approaches as a means to alleviate the need for massive data acquisition.
    In a tabular scenario where the Q-functions are collected across tasks, we model our learning problem as optimizing a higher order tensor structure.  
    Recognizing that close-related tasks may require similar actions, our proposed method imposes a low-rank condition on this aggregated Q-tensor. The rationale behind this approach to multi-task learning is that the low-rank structure enforces the notion of similarity, without the need to explicitly prescribe which tasks are similar, but inferring this information from a reduced amount of data simultaneously with the stochastic optimization of the Q-tensor. 
    The efficiency of our low-rank tensor approach to multi-task learning is demonstrated in two numerical experiments, first in a benchmark environment formed by a collection of inverted pendulums, and then into a practical scenario involving multiple wireless communication devices.

\end{abstract}


%%%%%%%%%%%%%%%%%%%%%%%%%%%%%%%%%%%%%%%%%%
\section{Introduction}
\label{S:intro}

Reinforcement Learning (RL) has emerged as a cornerstone of modern artificial intelligence, driving remarkable advances across diverse fields. One early demonstration of the potential of learning machines was the 1997 match between Deep Blue and Garry Kasparov, a milestone in machine learning. Since then, RL has evolved into an essential component of highly sophisticated systems, such as the actor-critic algorithms that underpin the continuous improvement of tools like ChatGPT \cite{achiam2023gpt}. However, despite these advances, RL remains largely confined to virtual environments. A key barrier to its application in real-world physical systems is the massive amount of data required for training, which becomes prohibitive when data must be acquired through real-world interactions.
To address this challenge, researchers have turned to meta-learning as a strategy to significantly reduce the data requirements needed to achieve target performance.

Meta-learning takes many forms \cite{vanschoren2019meta,hospedales2020meta}.
One of them incorporates Bayesian priors derived from similar past experiments, with their influence diminishing as new data becomes available \cite{finn2018probabilistic,rakelly2019efficient}. When prior experiments involve agents trained on different but related tasks, it takes the form of multi-task learning, which integrates data acquired at several scenarios into a common problem \cite{caruana1997multitask,ruder2017overview,zhang2021survey,cervino2021multiTSP,cervino2021multi}. Approaches in multi-task learning include enforcing proximity constraints \cite{kato2008multi,kato2009conic,koppel2017proximity,koppel2019parsimonious}, adaptively sampling tasks \cite{sharma2017learning}, weighting task contributions in deep learning \cite{sener2018multi}, and jointly optimizing tasks in support vector machines \cite{Evengiou2004regularized, zhang2010convex}. In the form of transfer learning, the focus turns to be on adapting previously learned models to new tasks %
\cite{torrey2010transfer}.

In this work, we consider exploiting the information across tasks by learning a low-rank tensor representation. When the tabular Q-function for each task is represented as a matrix, the collection of these matrices across tasks forms a three-dimensional tensor. Our model can also accommodate higher-order tensors to handle multi-dimensional states and actions expressed as vectors. 
The motivation for using low-rank tensors in the context of multi-task RL is threefold: i) low rank has been shown to be pervasive across real-world applications, including in RL tasks \cite{rozada2023matrix, rozada2024tensor, rozada2024tensorb}; ii) the problem of obtaining the factors of a low-rank tensor is computationally tractable since the multilinear structure facilitates the computation of gradients; and iii) while being nonlinear, low-rank tensor models provide some degree of interpretability. 

In the following sections, we formulate the problem by modeling the multi-task Q-function as a Q-tensor and present algorithms that learn from data while enforcing a low-rank structure dynamically. We then demonstrate the effectiveness of our approach through two examples. The first is a benchmark experiment involving the stabilization of a diverse set of inverted pendulums. The second applies low-rank RL to optimize the scheduling of transmissions in a multi-device wireless communication system.

%%%%%%%%%%%%%%%%%%%%%%%%%%%%%%%%%%%%%%%%%%

%%%%%%%%%%%%%%%%%%%%%%%%%%%%%%%%%%%%%%%%%%
\section{Problem formulation}
\label{S:problem}

In this work, we consider the problem of learning the state-action value function (a.k.a. the Q function) for multiple related tasks. To be formal, consider $M$ separate MDPs denoted by $\mathcal{M}_m(\mathcal{S},\mathcal{A},P_m,r_m,\gamma)$ with $m=1,...,M$. In the previous expression, $\mathcal{S}$ and $\mathcal{A}$ denote the state and action spaces (common to all MDPs), $P_m\in\mathbb{R}^{|\mathcal{S}|\times|\mathcal{S}||\mathcal{A}|}$ denotes the matrix of transition probabilities for the $m$-th MDP, $r_m:\mathcal{S}\times\mathcal{A}\to \mathbb{R}$ is the reward function for the $m$-th MDP, and $\gamma \in(0,1)$ denotes the discount factor for all MDPs.
Let $t\in\mathbb{N}\cup\left\{0\right\}$ denote a time index, and let $s_t\in\mathcal{S}$ and $a_t\in \mathcal{A}$ be random variables denoting the state and action at time $t$. In RL, the objective is to find a policy, i.e., a stationary conditional distribution, to select actions based on the current state to maximize the expected cumulative reward:
\begin{equation}\label{eqn_rl}
\max_{\pi_m}\mathbb{E}_{P_m}\left[\sum_{t=0}^\infty \gamma^t r_m(S_t,A_t) \mid \pi_m\right],
\end{equation}
where the expectation is taken with respect to the transition probabilities $P_m$ and the policy $\pi_m$. 
For any MDP, say the $m$-th one, one can define the Q function: 
\begin{equation}\label{eqn_q_star}
    Q^{\pi_m}_m(s,a) = 
\mathbb{E}_{P_m}\left[\sum_{t=0}^\infty \gamma^t r_m(s_t,a_t) \mid \pi_m, s_0=s, a_0=a\right].
\end{equation}
The goal in this case is to find either the optimal Q-function (i.e., the one associated with the policy that maximizes the reward) or a good one (by restricting the value function or the policies to belong to a particular class of functions). The former typically requires exploiting the Bellman optimality equations and, for large problems, iterative algorithms such as Q-learning \cite{watkins1992q} and its variations \cite{bradtke1996linear, mnih2015human} are common in the literature. In most cases, estimating the optimal Q-function is infeasible due to the large number of states, and a parametric approach is pursued. To be specific, let us consider the value function $Q_m(s,a;\bbtheta_m)$, with $\bbtheta_m$ being the parameters. The problem of value function estimation reduces to estimating the optimal value of $\bbtheta_m$. The parameters are then optimized using (stochastic/approximate/semi) gradient scheme:
\begin{equation}\label{eqn_q_parameters_upddate}
    \bbtheta_m^{(n+1)} = \bbtheta_m^{(n)} + \eta^{(n)} \hat{\nabla}\ell(\bbtheta_m^{(n)}),
\end{equation}
where $n$ is the iteration index, $\eta^{(n)}$ is a learning rate, and $\ell$ is a loss function. A popular choice, which is the one considered in this paper, is to sample trajectories from the MDP to collect transitions $\bbsigma_m^l=\{(a_l,s_l,r_l,s'_l)\}$ and define the loss as 
\begin{equation}\label{eqn_loss_for_q_parameters_upddate}
       \ell(\bbtheta_m)\!=\!\!\sum_{l=1}^{L_m}
         \big( r_l + \gamma \max_{a} Q_m\!(s'_l, a;\bbtheta_m) - Q_m(s_l, a_l;\bbtheta_m)\!\big)^{\!2}\!.
\end{equation}

If the similarities across tasks are neglected, 
then one can optimize the $M$ instances of the loss in \eqref{eqn_loss_for_q_parameters_upddate} separately, yielding $M$ Q-functions. A more reasonable approach, especially when the dimensionality is large or the number of samples is small, is to exploit those similarities. A principled way to do this is: a) to define a single optimization problem for the $M$ tasks where we combine the objectives of each of them, and b) to introduce constraints or regularizers promoting that either the Q-functions themselves or the parameters are similar and/or share some hyperparameters. Regarding a), in the case of \eqref{eqn_loss_for_q_parameters_upddate}, this simply entails defining $\bar{\ell}(\bbTheta)=\sum_{m=1}^M \lambda_m \ell(\bbtheta_m)$, with $\bbTheta=[\bbtheta_1,...,\bbtheta_M]$ and $\lambda_m$ being nonnegative coefficients.

The question remains on how to impose similarities across tasks. While policy-based methods achieve similar policies by constraining their Kullback–Leibler divergence or clipping the loss \cite{schulman2017proximal}, we propose an alternative approach tailored to jointly learning Q-functions associated with multiple tasks. 

Our method is  based on the following three modeling considerations. First, we leverage the fact that we work with finite state and action spaces to model the Q-function of each task as a matrix (or a tensor; see details in the next subsection). We then combine the $M$ Q-functions into \textit{a single tensor $\tenbQ$ with one additional mode that indexes the tasks}. Second, and crucially,\textit{ we postulate that the Q-tensor $\tenbQ$ has low rank} and recast the problem of learning the Q-function as the estimation of the tensor factors. Third, we develop schemes to compute the tensor factors jointly for the $M$ tasks. Unlike Deep Q-learning \cite{mnih2013playing}, our parameters $\theta_m$ do not represent the weights of a neural network. Moreover,  we do not  maintain an explicit separate set of parameters $\theta_m$ for each task. Instead, we combine the parameters corresponding to all tasks into a set of low-rank tensor factors and use them to form a parametric model $\bbTheta=\mathrm{factors}(\tenbQ)$ that represents all tasks jointly. Accordingly, the similarity between tasks does not need to be enforced by proximal constraints on the parameters; rather, it is inherently presumed in the low-rank model.

\subsection{Tensors and their application to Q-function estimation}

%%%%%%%%%%%%%%%%%%%%%%%%%%%%%%%%%%%%%%%%%%%%%%%%%%%%%%%%%%%%

\begin{table*}

\begin{subequations}
\begin{align}
    \bar{\ell}(\bbQ_1,\bbQ_2,\bbQ_3)&=\sum_{m=1}^{M} \!\!\lambda_m   \sum_{l=1}^{L_m} (r_l + \gamma \max_{i_a} \tenbQ(i_{s'_l}, i_a, m) - \tenbQ(i_{s_l}, i_{a_l}, m))^2~\text{s.to:}\; 
    \tenbQ(i_1,i_2,i_3)=\sum_{k=1}^K[\bbQ_1]_{i_1,k}[\bbQ_2]_{i_2,k}[\bbQ_3]_{i_1,k}; \label{eq_cost_multitasktensor_with_constraint}\\
    \bar{\ell}(\bbQ_1,\bbQ_2,\bbQ_3)&=\sum_{m=1}^{M} \lambda_m\!\!  \sum_{l=1}^{L_m} \big( r_l + \gamma \max_{i_a} \sum_{k=1}^K[\bbQ_1]_{i_{s'_l},k}[\bbQ_2]_{{i_a},k}[\bbQ_3]_{m,k}- \sum_{k=1}^K[\bbQ_1]_{i_{s_l},k}[\bbQ_2]_{i_{a_l},k}[\bbQ_3]_{m,k}\big)^2.\label{eq_cost_multitasktensor_without_constraint}
\end{align}
\end{subequations}

\vspace{-.2cm}
\end{table*}
%%%%%%%%%%%%%%%%%%%%%%%%%%%%%%%%%%%%%%%%%%%%%%%%%%%%%%%%%%%%
%%%%%%%%%%%%%%%%%%%%%%%%%%%%%%%%%%%%%%%%%%%%%%%%%%%%%%%%%%%%
\begin{table*}
\begin{subequations}\label{eq_semigradients}
\begin{align}\label{eq_semigradientQstate}
    \!\!\!\!\!\! [\hat{\nabla}_{\bbQ_1}^{\bbsigma_{m}^t}\!\bar \ell(\bbQ_1,\bbQ_2,\bbQ_3)]_{i_s\!,k}\!=2\lambda_m \!\! &\left( r_t + \gamma \max_{i_a} \sum_{k=1}^K[\bbQ_1]_{i_{\!s'_t}\!,k}[\bbQ_2]_{{i_a}\!,k}[\bbQ_3]_{m,k}-\! \sum_{k=1}^K[\bbQ_1]_{i_{\!s_t},k}[\bbQ_2]_{i_{a_t}\!,k}[\bbQ_3]_{m,k}\right) \!\!\! \left( [\bbQ_2]_{a_t\!,k}\,[\bbQ_3]_{m,k}\,\mathbb{I}_{\big\{i_s=i_{s_t}\big\}} \right);
\end{align}
\begin{align}\label{eq_semigradientQaction}
    \!\!\!\! [\hat{\nabla}_{\bbQ_2}^{\bbsigma_{m}^t}\!\bar \ell(\bbQ_1,\bbQ_2,\bbQ_3)]_{i_a\!,k}\!=2\lambda_m \!\! &\left( r_t + \gamma \max_{i_a} \sum_{k=1}^K[\bbQ_1]_{i_{s'_t}\!,k}[\bbQ_2]_{{i_a}\!,k}[\bbQ_3]_{m,k}-\! \sum_{k=1}^K[\bbQ_1]_{i_{\!s_t},k}[\bbQ_2]_{i_{a_t}\!,k}[\bbQ_3]_{m,k}\right) \!\!\! \left(  [\bbQ_1]_{s_t\!,k}\,[\bbQ_3]_{m,k}\,\mathbb{I}_{\big\{i_a=i_{a_t}\big\}} \right);
\end{align}
\begin{align}\label{eq_semigradientQtask}
    \!\!\!\! [\hat{\nabla}_{\bbQ_3}^{\bbsigma_{m}^t}\!\bar \ell(\bbQ_1,\bbQ_2,\bbQ_3)]_{m,k}\!=2\lambda_m\!\! &\left( r_t + \gamma \max_{i_a} \sum_{k=1}^K[\bbQ_1]_{i_{s'_t}\!,k}[\bbQ_2]_{{i_a}\!,k}[\bbQ_3]_{m,k}-\! \sum_{k=1}^K[\bbQ_1]_{i_{\!s_t},k}[\bbQ_2]_{i_{a_t}\!,k}[\bbQ_3]_{m,k}\right) \!\!\! \left(  [\bbQ_1]_{s_t\!,k}\,[\bbQ_2]_{a_t\!,k}\,\mathbb{I}_{\Big\{\substack{i_s= i_{s_t} \\ i_a = i_{a_t}} \Big\}} \right)\!\!.
\end{align}
\end{subequations}
\vspace{-.2cm}
\end{table*}
%%%%%%%%%%%%%%%%%%%%%%%%%%%%%%%%%%%%%%%%%%%%%%%%%%%%%%%%%%%%

As mentioned in the previous section, our approach is to consider a tensor $\tenbQ$ that collects the Q-functions of the $M$ tasks.
When $\ccalS$ and $\ccalA$ have a single dimension, each task has a Q-matrix, and our tensor has dimensions $\tenbQ\in\reals^{|\ccalS|\times |\ccalA|\times M}$. With this notation, one entry of the tensor, say $\tenbQ(i_s,i_a,m)$, represents $\mathbb{E}_{P_m}\left[\sum_{t=0}^\infty \gamma^t r_m(s_t,a_t) \mid \pi_m, s_0=s_{i_s}, a_0=a_{i_a}\right]$, with $s_{i_s}$ denoting the state indexed by $i_s$ and $a_{i_a}$ the action indexed by $i_a$.

We now consider that the tensor $\tenbQ$ has low rank. Multiple definitions of rank exist, and the focus of this paper is on the PARAFAC rank \cite{sidiropoulos2017tensor}. Under this definition, the rank of a tensor can be defined as the minimum number of \emph{rank-1} tensors that need to be added to recover the entire tensor. A rank-1 tensor of dimension $D$ is simply the outer product of $D$ vectors:
\begin{align}\nonumber
    \tenbQ_{rank-1}=\bbq_1\circledcirc...\circledcirc\bbq_D~ \Leftrightarrow~\tenbQ_{rank-1}(i_1,...,i_D)=\prod_{d=1}^D[\bbq_d]_{i_d}
\end{align}
Hence, a tensor with rank $K$ is simply 
\[
\tenbQ = \sum_{k=1}^K [\bbQ_1]_{:,k} \circledcirc ... \circledcirc [\bbQ_D]_{:,k} ~ \Leftrightarrow~\tenbQ(i_1,...,i_D)=\sum_{k=1}^K\prod_{d=1}^D[\bbQ_d]_{i_d,k},
\]
with $\bbQ_d=[\bbq_d^1,...,\bbq_d^K]$ being the so-called factor matrices, whose $k$-th column $[\bbQ_d]_{:,k}$ collects the vector associated with the $k$-th rank-1 tensor for the $d$-th dimension (mode) of the tensor. Note that by imposing low rank, the number of degrees of freedom in $\tenbQ$ shifts from $|\ccalS| |\ccalA| M$ to $(|\ccalS|+|\ccalA|+M)K$, which entails significant savings both in terms of computations and sample complexity.

As mentioned in Section \ref{S:intro} one advantage of of considering tensor formulations is their interpretability. To elaborate on this point,
consider our 3-dimensional tensor $\tenbQ$ and let $\bbQ_1\in\reals^{|\ccalS|\times K}$ be the factor matrix associated with the states, $\bbQ_2\in\reals^{|\ccalA|\times K}$ be the factor matrix associated with the actions, and $\bbQ_3\in\reals^{M\times K}$ be the factor matrix associated with the tasks. Using the two first matrices, we form the $|\ccalS|\times |\ccalA|$ action-space matrix associated with the $k$-th rank-1 component of the tensor as
$\tbQ^k = [\bbQ_1]_{:,k}([\bbQ_2]_{:,k})^\top.$    
Then, it holds that we can understand the $|\ccalS|\times |\ccalA|$ Q-matrix associated with the $m$-th task as 
\begin{align} \label{eq_slicesQm}
\bar{\bbQ}_m = \sum_{k=1}^K [\bbQ_3]_{m,k} \tbQ^k=\sum_{k=1}^K [\bbQ_3]_{m,k} \left([\bbQ_1]_{:,k}([\bbQ_2]_{:,k})^\top\right).
\end{align}
In words, the low-rank tensor decomposition provides a latent representation of the Q-matrix into a $K$-dimensional space [c.f. \eqref{eq_slicesQm}], and then synthesizes the Q-matrix associated with the $m$-th task as the inner product of that latent representation with the $K$ coefficients encoded in the $m$-th row of the factor matrix $\bbQ_3$. 

As a result of this low-dimensional embedding we can learn the parameters of our tensor model  $\mathbf \Theta=\{\bbQ_1,\bbQ_2,\bbQ_3\}$ jointly using information (samples) from all the tasks, with only one  hyperparameter being the rank $K$.
Hence, the next subsection introduces the problem of learning $\mathbf \Theta$  formally, but a remark is in order first.

\begin{remark}
    In most practical scenarios, both state and action spaces have multiple dimensions. In this case, rather than considering $\tenbQ$ to have 3 modes, it is more natural to model $\tenbQ$ as a $D_{\mathcal{S}}+D_{\mathcal{A}}+1$ dimensional tensor, with $D_{\mathcal{S}}$ denoting the number of dimensions of the state space, $D_{\mathcal{A}}$ denoting the number of dimensions of the action space, and the additional dimension indexing the tasks. Low-rank tensor models are even better motivated in this case, with the computational and sample complexity savings being larger for the $D_{\mathcal{S}}+D_{\mathcal{A}}+1$ dimensional case. For simplicity, this conference paper will model $\tenbQ$ as a tensor with only 3 modes, but higher-order modeling will be explored in the journal version of this paper. 
\end{remark}

\subsection{Multitask low-rank tensor formulation}

To formulate our multitask low-rank Q-tensor estimation problem, we first consider having access to sampled transitions from all the MDPs, and denote these sets as $\ccalT_1=\{\bbsigma_1^{l}\}_{l=1}^{L_1},...,\ccalT_M=\{\bbsigma_M^{l}\}_{l=1}^{L_M}~\text{with}\;\bbsigma_1^{l}=(s_l,a_l,r_l,s'_l)$. 
With these sets, and using the weights $\{\lambda_m>0\}_{m=1}^M$, we define the single multitask cost using the expression in \eqref{eq_cost_multitasktensor_with_constraint} located at the top of the page, where we recall that $i_{s}$ and $i_{a}$ denote the index associated with state $s$ and action $a$, respectively. Alternatively, we can replace the constraint directly into the objective, yielding the expression in \eqref{eq_cost_multitasktensor_without_constraint}.
Leveraging \eqref{eq_cost_multitasktensor_without_constraint}, the optimal factors can then be found as
\begin{equation}
    \{\bbQ_1^*,\bbQ_2^*,\bbQ_3^*\}=\arg\min_{\bbQ_1,\bbQ_2,\bbQ_3} \bar{\ell}(\bbQ_1,\bbQ_2,\bbQ_3).
\end{equation}
The above optimization is complex since: i) the length and number of trajectories may be large, rendering the computation of the gradient costly; ii) due to the multilinearity, the low-rank tensor decomposition is not convex; and iii) the presence of the max operator in \eqref{eq_cost_multitasktensor_without_constraint}. Section \ref{S:method} discusses an effective method to approximate $\{\bbQ_1^*,\bbQ_2^*,\bbQ_3^*\}$.

%%%%%%%%%%%%%%%%%%%%%%%%%%%%%%%%%%%%%%%%%%

%%%%%%%%%%%%%%%%%%%%%%%%%%%%%%%%%%%%%%%%%%
\section{Proposed algorithm}
\label{S:method}
This section describes an online algorithm to find the parameters $\mathbf \Theta=\{\bbQ_1,\bbQ_2,\bbQ_3\}$ collecting the factors that define the multi-task low-rank tensor $\tenbQ\in\reals^{|\ccalS|\times |\ccalA| \times M}$. Important features of our algorithm are: i) it is a stochastic scheme that utilizes samples of the trajectory sets; ii) it implements a block coordinate descent approach across factors; and iii) it uses a semi-gradient scheme to deal with the max operator. The latter is a widely-adopted approach in value-based RL due to its simplicity and good practical performance \cite{geist2013algorithmic}.

In particular, we run episodic trajectories of $T$ time-steps, and after sampling the $t$-th transition, say $\bbsigma_{m}^t$, we update the factors $\{\bbQ_1,\bbQ_2,\bbQ_3\}$ using an iterative algorithm as follows
\begin{subequations}\label{eq_stochasticsemigradientupdates}
\begin{align}
    \!\!\bbQ_1^{(n+1)}&=\bbQ_1^{(n)}-\eta^{(n)}\hat{\nabla}_{\bbQ_1}^{\bbsigma_{m}^t}\bar{\ell}(\bbQ_1^{(n)},\bbQ_2^{(n)},\bbQ_3^{(n)})
    \label{eq_stochasticsemigradientupdate_states}\\
    \!\!\bbQ_2^{(n+1)}&=\bbQ_2^{(n)}-\eta^{(n)}\hat{\nabla}_{\bbQ_2}^{\bbsigma_{m}^t}\bar{\ell}(\bbQ_1^{(n)},\bbQ_2^{(n)},\bbQ_3^{(n)}) \label{eq_stochasticsemigradientupdate_actions}\\
    \!\!\!\![\bbQ_3^{(n+1)}]_{m,:}&= \! [\bbQ_3^{(n)}]_{m,:} \!\! -\eta^{(n)}[\hat{\nabla}_{\bbQ_3}^{\bbsigma_{m}^t}\bar{\ell}(\bbQ_1^{(n)},\bbQ_2^{(n)},\bbQ_3^{(n)})]_{m,:} \label{eq_stochasticsemigradientupdate_tasks}
\end{align}
\end{subequations}
where $n$ is an iteration index, $\eta^{(n)}$ is the learning rate, and the expressions for the semi-gradients are given in \eqref{eq_semigradients}, which is located at the top of the page. In those expressions, $\mathbb{I}_{\{\cdot\}}$ is the indicator function (one if the Boolean argument is true and zero otherwise). To ensure exploration, the semi-gradients in \eqref{eq_semigradients} will implement a maximization over the action space with probability $(1-\varepsilon)$ and will take a random action with probability $\varepsilon$. To help readability, the pseudocode for our scheme, denoted as S-TLR-Q, is provided in Algorithm \ref{Alg:Algorithm_Update_TensorMultitask}.

\begin{algorithm} \footnotesize
\caption{Stochastic Tensor Low-Rank Q-Est. (S-TLR-Q)}            \label{Alg:Algorithm_Update_TensorMultitask} \vspace{0.01cm}
    \SetKwInput{Input}{Inputs}
    \Input{$\eta^{(n)}$, $N$, $K$, $T$, $\{\lambda_m\}_{m=1}^M$, $\varepsilon$.}
    \SetAlgoLined
    \textbf{Initialization:} $[\bbQ_1^{(0)},\bbQ_2^{(0)},\bbQ_3^{(0)}] \sim \mathrm{Uniform}(0,1)$ and $n=1$.\\ 
    \While{true}{
        \For{$m = 1, 2, \ldots, M$}{
            \For{$t = 1, 2, \ldots, T$}{
                Sample transition $\bbsigma_m^t$ \\
                \BlankLine
                \BlankLine
                Update $\bbQ_1^{(n+1)}$ using $\bbsigma_m^t$, \eqref{eq_stochasticsemigradientupdate_states}, \eqref{eq_semigradientQstate}, $\eta^{(n)}$ and $\varepsilon$.\\
                Update $\bbQ_2^{(n+1)}$ using $\bbsigma_m^l$, \eqref{eq_stochasticsemigradientupdate_actions}, \eqref{eq_semigradientQaction}, $\eta^{(n)}$, and $\varepsilon$.\\
                Update $[\bbQ_m^{(n+1)}]_{m,:}$ using $\bbsigma_m^l$, \eqref{eq_stochasticsemigradientupdate_tasks}, \eqref{eq_semigradientQtask}, $\eta^{(n)}$, and $\varepsilon$.
                \BlankLine
                \BlankLine
                \If{$n=N$}{
                    \textbf{break}
                }
                Update $n=n+1$
            }
        }
    }
    \SetKwInOut{Output}{Outputs}
    \Output{$\bbQ_1^{(N)},\, \bbQ_2^{(N)},\, \bbQ_3^{(N)}$}
\end{algorithm}

\begin{figure*}[t]
    \centering
    \begin{minipage}[c]{.23\textwidth}
        \includegraphics[width=\textwidth]{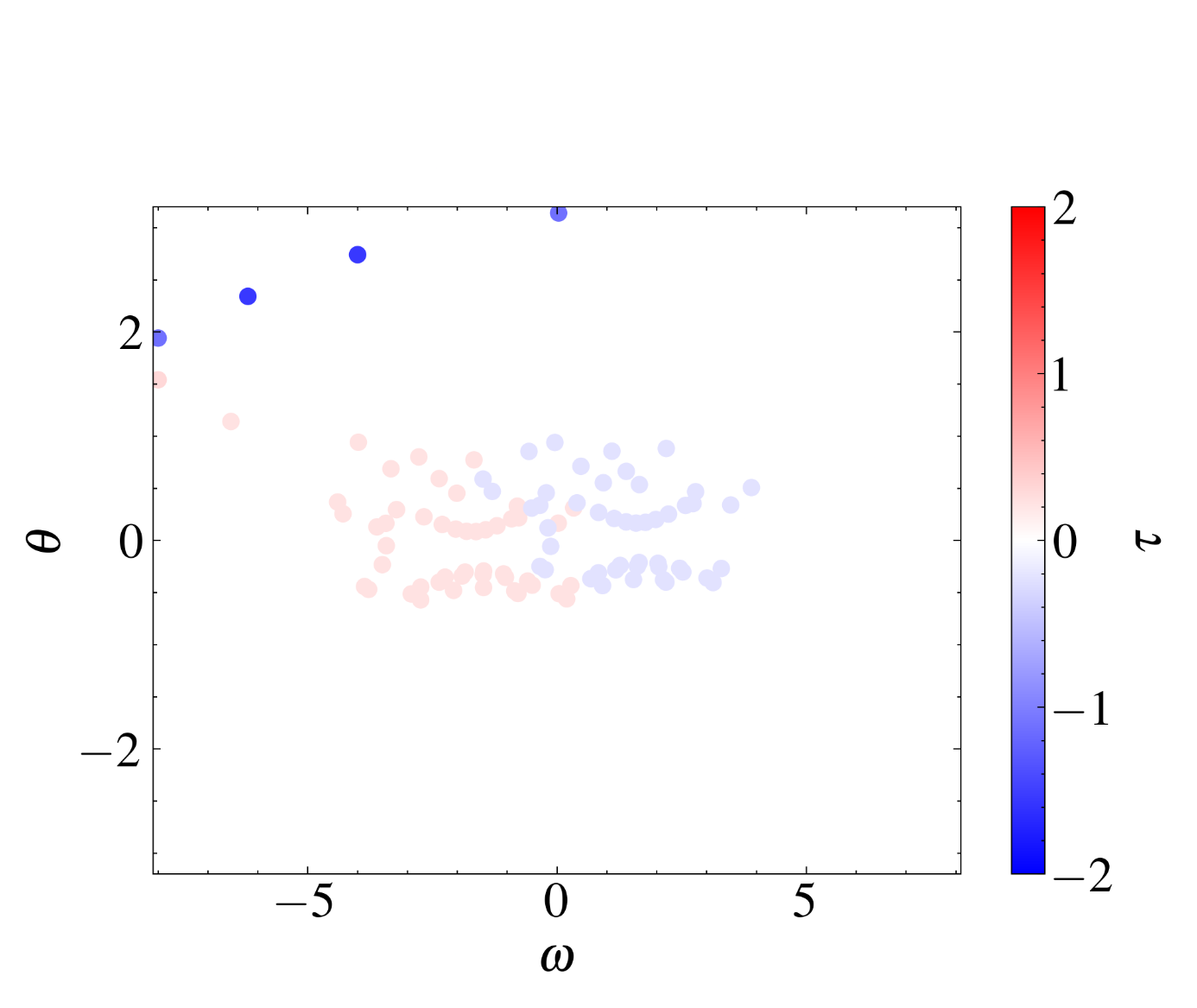}
        %\centering{\footnotesize (a) Pendulum with  $g_1=0.01$, and $d_1=1.0$.}
    \end{minipage}
    \hspace{0.1cm}
    \begin{minipage}[c]{.23\textwidth}
        \includegraphics[width=\textwidth]{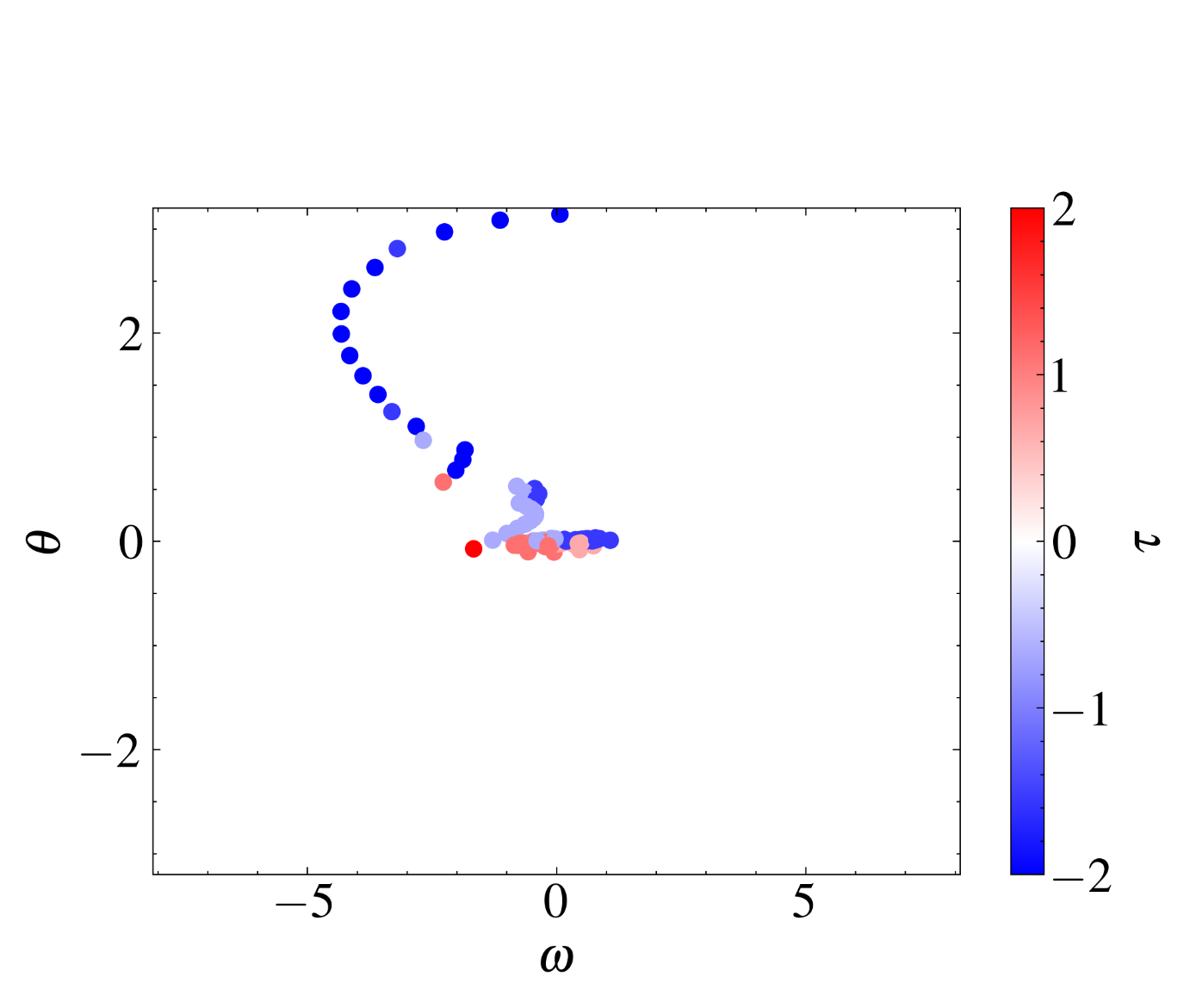}
        %\centering{\footnotesize (b) Pendulum with  $g_4=1.0\;\;$, and $d_4=0.5$.}
    \end{minipage}
    \hspace{0.1cm}
    \begin{minipage}[c]{.23\textwidth}
        \vspace{0.2cm}\includegraphics[width=\textwidth]{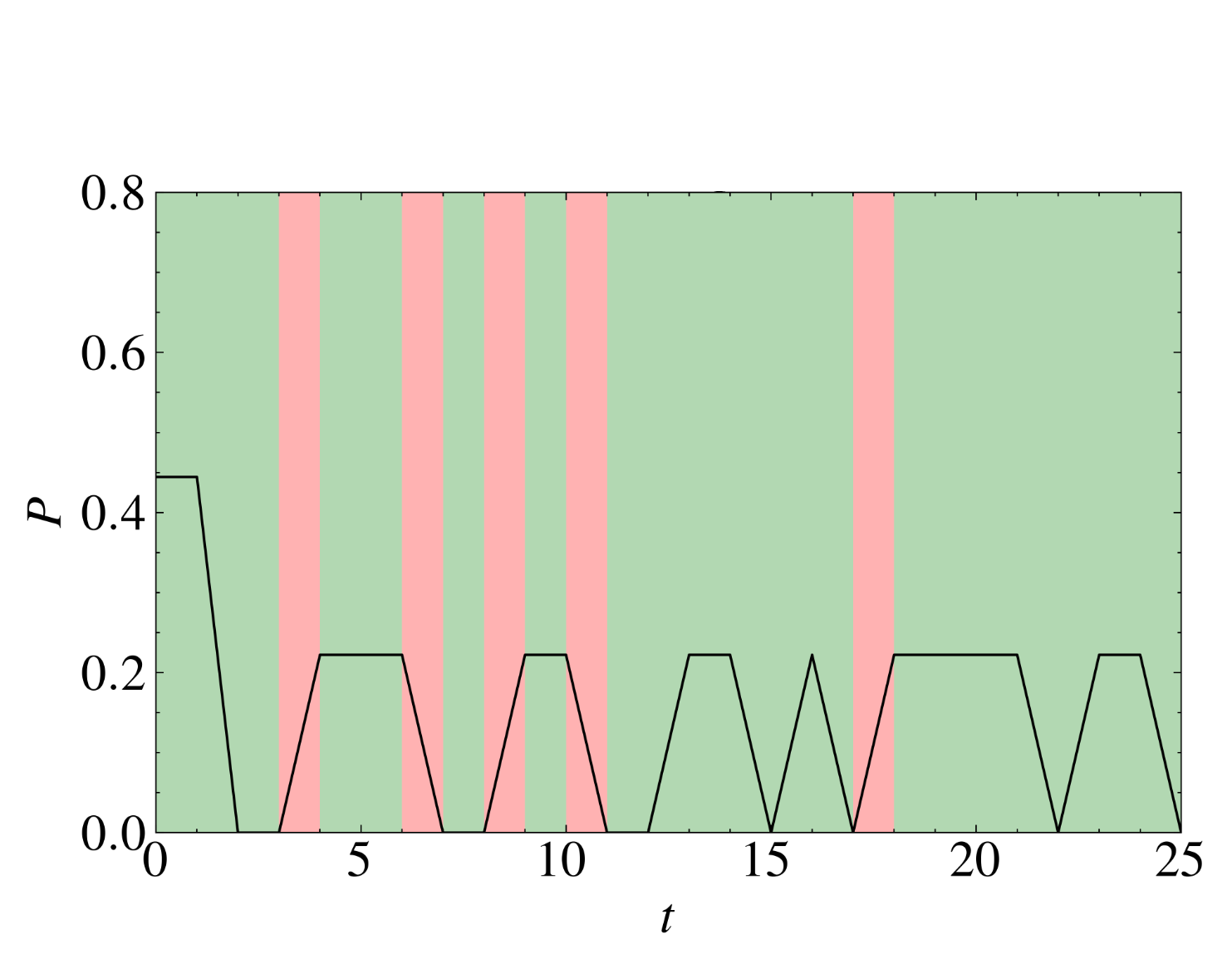}
        %\centering{\footnotesize (c) Wireless with $a_1=1.0$, $b_1=0.5$, $p_{a_1}=p_{b_1}=0.2$}.
    \end{minipage}
    \hspace{0.1cm}
    \begin{minipage}[c]{.23\textwidth}
        \vspace{0.2cm}\includegraphics[width=\textwidth]{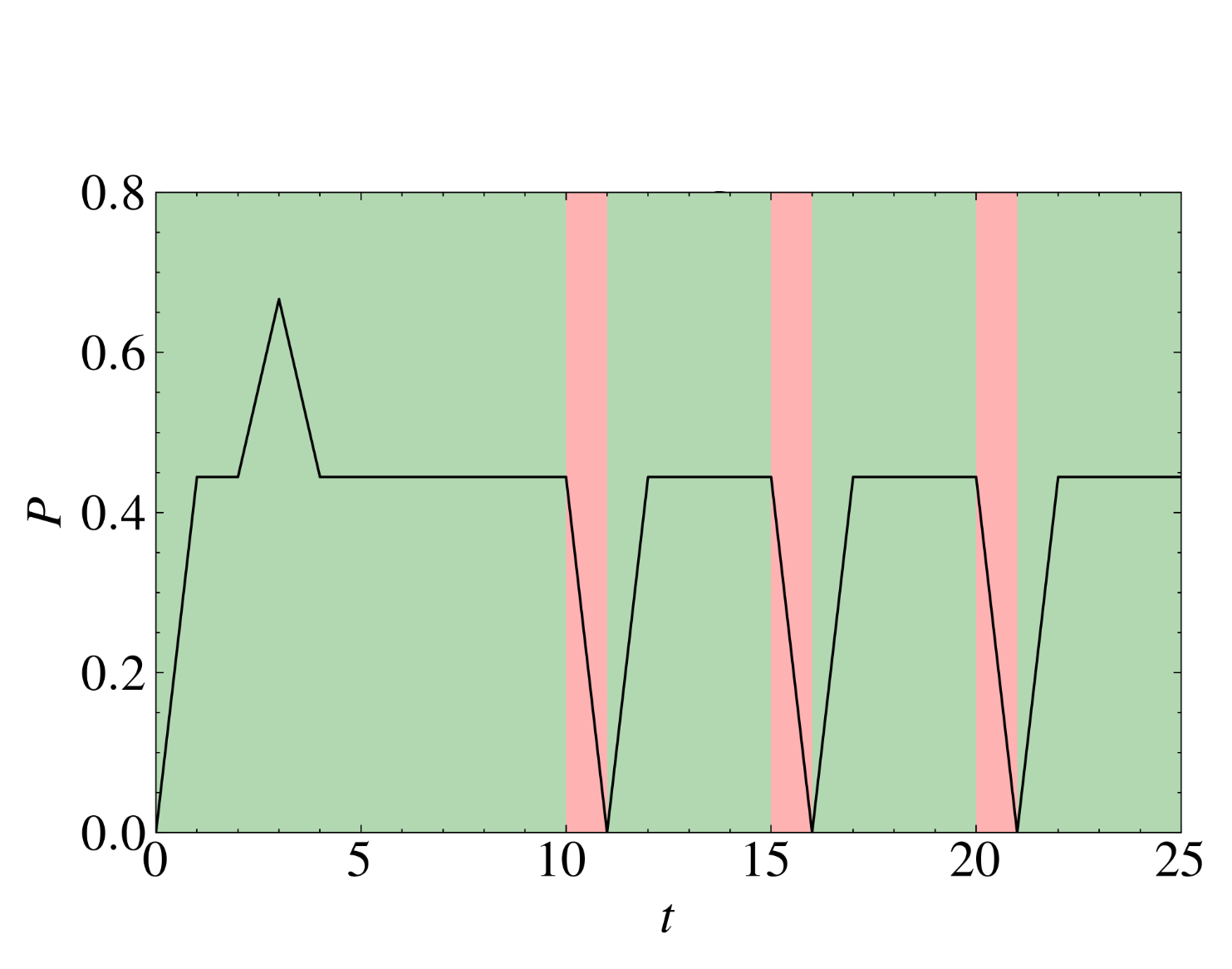}
        %\centering{\footnotesize(d) Wireless with $a_4=2.0$, $b_4=3.0$, $p_{a_4}=p_{b_4}=0.8$}.
    \end{minipage}
    \vspace{-.1cm}
    \caption{
    Trajectories sampled using the estimated optimal policy learned with \textbf{LR-Q} for different scenarios.
    From left to right: the pendulum problem with $(g_1=0.01,d_1=1.0)$ (1st panel) and $(g_1=0.01,d_1=1.0)$ (2nd panel); and the wireless setup with  $\alpha_1=1.0$, $b_1=0.5$, $p_{\alpha_1}=p_{b_1}=0.2$ (3rd panel) and $\alpha_4=2.0$, $b_4=3.0$, $p_{\alpha_4}=p_{b_4}=0.8$ (4th panel). While the optimal trajectories differ across the scenarios, the estimated policies show structural similarities.\vspace{-.35cm}
    }
\label{f:setups}
\end{figure*}

It is important to note that, when a transition of the task $m$ is sampled, our algorithm updates: i) the embeddings for the states and actions observed during the trajectory [cf. \eqref{eq_stochasticsemigradientupdate_states}, \eqref{eq_stochasticsemigradientupdate_actions}, \eqref{eq_semigradientQstate} and \eqref{eq_semigradientQaction}] and ii) the coefficients associated with task $m$ [cf. \eqref{eq_stochasticsemigradientupdate_tasks} and \eqref{eq_semigradientQtask}]. The operation of the algorithm clearly illustrates the benefits of the joint (multi-task) scheme, since the updates associated with trajectories of task $m$ impact the value functions of all the other tasks, accelerating convergence and reducing sample complexity relative to the separate scheme that estimates $M$ Q-functions separately.

%%%%%%%%%%%%%%%%%%%%%%%%%%%%%%%%%%%%%%%%%%

%%%%%%%%%%%%%%%%%%%%%%%%%%%%%%%%%%%%%%%%%%
\section{Numerical Analysis}
\label{S:results}

This section demonstrates that \textbf{S-TLR-Q} benefits from concurrent learning across tasks in two scenarios: the classical pendulum control problem \cite{brockman2016openai} and a wireless communications setup. We compare \textbf{S-TLR-Q} against two baselines: \textbf{LR-Q}, which learns Q-values for each task independently, and \textbf{C-LR-Q}, which learns a shared Q-value representation for all tasks. Both approaches leverage the methods in \cite{rozada2024tensor}. Additional details are provided in the code repository \cite{rozada2024code}.

\noindent \textbf{Pendulum control.} We consider the classic pendulum control problem, where the goal is to stabilize an underactuated pendulum in the upright position. The system consists of a pole with length $d$ and mass $g$ attached to a pivot, with dynamics governed by the angle $\theta$ (measured from the upright position) and angular velocity $\omega$. The agent applies a bounded torque $\tau$ at each time step to control the motion of the pendulum, with dynamics dictated by the equations of motion, incorporating gravity and friction. The system state is given by $\theta$ and $\omega$, and the reward penalizes deviations from the upright position and excessive torque, encouraging efficient stabilization. 

We evaluate $M=4$ tasks,  characterized by different masses $g_m$ and lengths $d_m$. Specifically, we consider $\bbg=[0.01, 0.1, 0.5,$ $ 1.0]$ and $\bbd=[1.0, 1.0, 0.5, 0.5]$. The tasks induce distinct policies but share structural similarities exploitable via our low-rank model. To illustrate this, we compare two tasks where we aim to equilibrate a low mass and a large mass, respectively. Optimal policies are independently estimated using \textbf{LR-Q}. Fig.~\ref{f:setups}-a shows a trajectory for the low-mass task, while Fig.~\ref{f:setups}-b depicts the large-mass task. The low-mass task achieves the upright position more quickly, while the large-mass task requires greater torque to maintain the upright position. Despite these differences, both policies exhibit a similar structure: positive torque $\tau$ is applied for negative angular velocity $\omega$, and vice versa. This structural similarity underscores the suitability of low-rank techniques for this problem.

We evaluate \textbf{S-TLR-Q} against \textbf{LR-Q} and \textbf{C-LR-Q} by running $2,000$ episodes for each task, with $T\!=\!100$ time steps per episode. After every transition, \textbf{S-TLR-Q}, \textbf{C-LR-Q}, and \textbf{LR-Q} update their respective parameters. Note that \textbf{LR-Q} updates the parameters of separate models for each task. To have a fair comparison in terms of computational budget, for every transition in task $m$, \textbf{LR-Q} performs $M$ updates to the parameters of the corresponding model. 

\begin{figure*}[t]
    \centering
    \begin{minipage}[c]{.48\textwidth}
        \includegraphics[width=\textwidth]{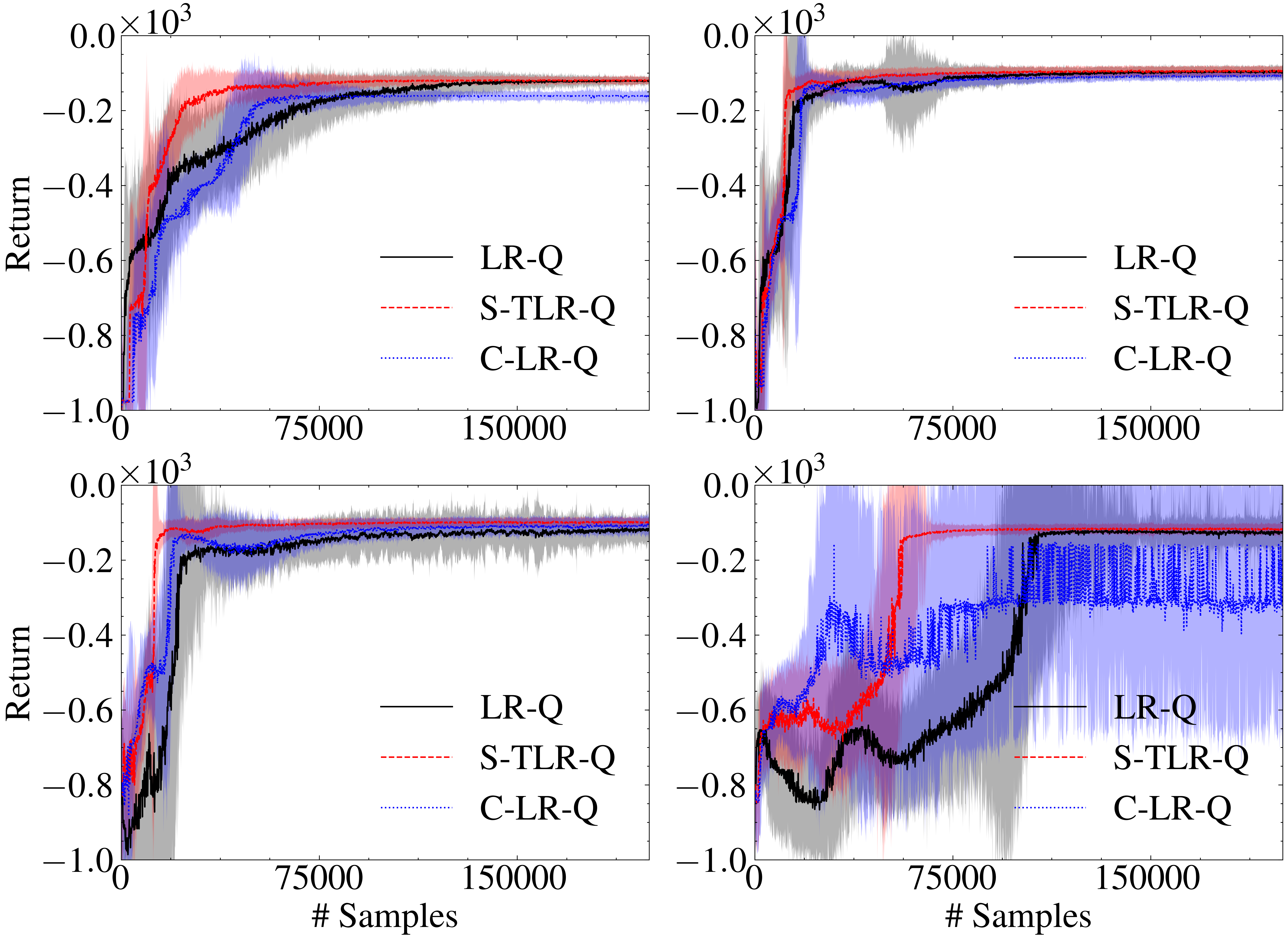}
        \centering{\footnotesize (a) Performance in the pendulum problem for $M=4$ tasks.}
    \end{minipage}
    \hspace{0.3cm}
    \begin{minipage}[c]{.48\textwidth}
        \includegraphics[width=\textwidth]{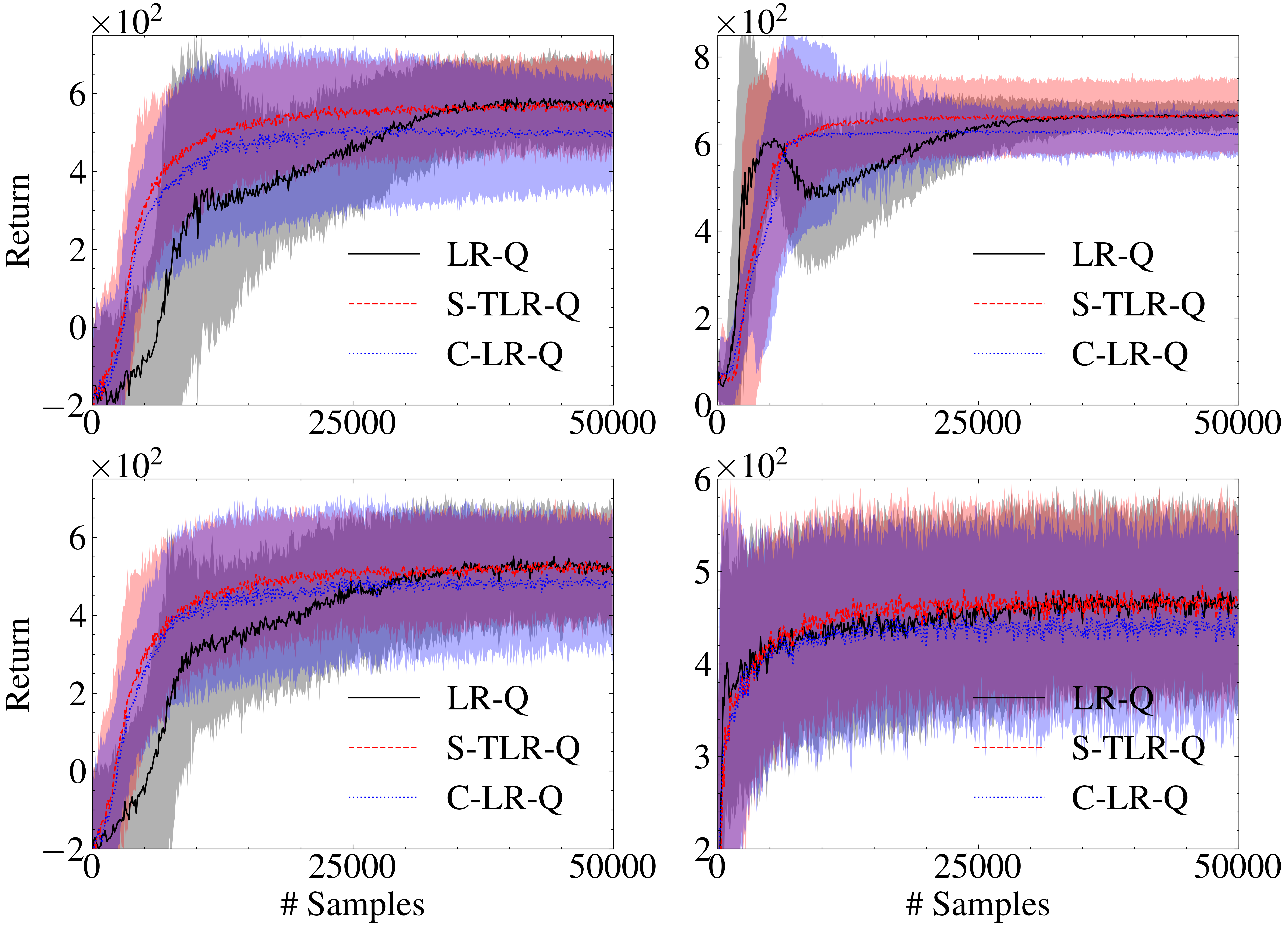}
        \centering{\footnotesize (b) Performance in the wireless problem for $M=4$ tasks.}
    \end{minipage}
    \caption{
    Performance of \textbf{S-TLR-Q} in (a) the pendulum problem and (b) the wireless setup, measured in terms of average return over $100$ experiments. \textbf{S-TLR-Q} requires fewer samples to converge than \textbf{LR-Q} and achieves higher returns than \textbf{C-LR-Q} consistently across all tasks.\vspace{-.2cm}
    }
\label{f:results}
\end{figure*}

Performance is measured in terms of the cumulative reward achieved in a test episode conducted after sampling transitions. Results are averaged across $100$ independent experiments. Fig.~\ref{f:results}-a presents the results for all tasks. As expected, \textbf{S-TLR-Q} and \textbf{C-LR-Q} converge faster and require fewer samples compared to \textbf{LR-Q}. However, \textbf{C-LR-Q} converges to a suboptimal solution, as it attempts to address all tasks using the same representation. In contrast, \textbf{S-TLR-Q} achieves performance comparable to \textbf{LR-Q}, which learns a separate model for each task. Importantly, \textbf{S-TLR-Q} leverages the low-rank structure to achieve these results with fewer samples, demonstrating its effectiveness in multi-task learning.

\vspace{.05cm}
\noindent \textbf{Wireless device.} We simulate an opportunistic multiple-access wireless setup involving a single agent with a queue and a battery transmitting packets to an access point over an orthogonal channel. At each time step, $\alpha$ packets arrive with probability $p_\alpha$, and the agent opportunistically accesses the channel, which may be occupied. The system state includes (a) the fading level and channel occupancy, (b) the battery energy level, and (c) the queue size. The agent selects the transmission power, with the transmission rate determined by Shannon's capacity formula. If the channel is occupied, packet loss is $100\%$. The battery harvests $b$ units of energy per time step with probability $p_b$. The reward balances positively weighted battery level and negatively weighted queue size, encouraging a trade-off between transmission throughput and battery efficiency. We evaluate $M=4$ tasks, each defined by distinct parameters: packet arrivals $\alpha_m$, harvested energy $b_m$, and arrival probabilities $p_{a_m}$ and $p_{b_m}$. Specifically, we consider $\bbalpha=[1.0, 1.0, 1.0, 2.0]$, $\bbb =[0.5, 0.5, 0.5, 3.0]$, $\bbp_{\alpha}=[0.2, 0.2, 0.5, 0.8]$, and $\bbp_b=[0.2, 0.5, 0.5, 0.8]$.

Similar to the pendulum setup, the tasks in the wireless setup induce distinct policies but share structural similarities. To illustrate, we compare two tasks: one with low packet arrivals and energy harvesting, and another with high packet arrivals and energy harvesting. Optimal policies are independently estimated using \textbf{LR-Q}. Fig.~\ref{f:setups}-c shows a trajectory for the low-packet task, where the agent transmits with low power on available channels and refrains from transmitting when channels are occupied. Fig.~\ref{f:setups}-d depicts the high-packet task, where the agent transmits more frequently with higher power, even under contention. Despite these differences, both policies exhibit a similar structure: the agent avoids transmission on occupied channels and scales transmission power based on packet size.

We evaluate \textbf{S-TLR-Q} against \textbf{LR-Q} and \textbf{C-LR-Q} in the wireless setup by running $500$ episodes per task, each consisting of $T=200$ time steps. As before, performance is assessed using the cumulative reward averaged over $100$ independent experiments. Fig.~\ref{f:results}-b presents the results. Once again, \textbf{S-TLR-Q} and \textbf{C-LR-Q} converge faster and require fewer samples compared to \textbf{LR-Q}. However, \textbf{C-LR-Q} converges to a suboptimal solution, while \textbf{S-TLR-Q} matches the performance of \textbf{LR-Q}. Importantly, \textbf{S-TLR-Q} consistently leverages the low-rank structure to significantly reduce the  required number of samples, demonstrating its effectiveness in multi-task  learning.

%%%%%%%%%%%%%%%%%%%%%%%%%%%%%%%%%%%%%%%%%%

%%%%%%%%%%%%%%%%%%%%%%%%%%%%%%%%%%%%%%%%%%
\section{Conclusion}
\label{S:conclusion}

We proposed a parametric model for multitask learning for the case of finite but prohibitively large dimensional spaces of states and actions. Our model collects Q-functions of multiple tasks into a common tensor. Similarities between the tasks are embedded by imposing a low-rank structure. Then, we derived the S-TRL-Q algorithm, which uses data from all tasks to learn the tensor factors. As we control the size of these factors by adjusting the rank, we can significantly reduce the number of parameters to be estimated, leading to higher performance in data-scarce regimes.  We run two experiments in support of this claim, with the goal of equilibrating a collection of inverted pendulums and maximizing the throughput of wireless devices, respectively. In both cases, S-TRL-Q obtains higher rewards than comparable methods, those that disregard differences or optimize tasks separately, especially in the earlier stages of training where the data is limited.

\balance
%%%%%%%%%%%%%%%%%%%%%%%%%%%%%%%%%%%%%%%%%%
\bibliographystyle{ieeetr}
\bibliography{references}

\begin{thebibliography}{10}

\bibitem{achiam2023gpt}
J.~Achiam, S.~Adler, S.~Agarwal, L.~Ahmad, I.~Akkaya, F.~L. Aleman, D.~Almeida, J.~Altenschmidt, S.~Altman, S.~Anadkat, {\em et~al.}, ``Gpt-4 technical report,'' {\em arXiv preprint arXiv:2303.08774}, 2023.

\bibitem{vanschoren2019meta}
J.~Vanschoren, ``Meta-learning,'' in {\em Automated Machine Learning}, pp.~35--61, Springer, Cham, 2019.

\bibitem{hospedales2020meta}
T.~Hospedales, A.~Antoniou, P.~Micaelli, and A.~Storkey, ``Meta-learning in neural networks: A survey,'' {\em arXiv preprint arXiv:2004.05439}, 2020.

\bibitem{finn2018probabilistic}
C.~Finn, K.~Xu, and S.~Levine, ``Probabilistic model-agnostic meta-learning,'' {\em Advances in Neural Info. Process. Syst.}, vol.~31, 2018.

\bibitem{rakelly2019efficient}
K.~Rakelly, A.~Zhou, C.~Finn, S.~Levine, and D.~Quillen, ``Efficient off-policy meta-reinforcement learning via probabilistic context variables,'' in {\em Intl. Conf. on Machine Learning (ICML)}, pp.~5331--5340, 2019.

\bibitem{caruana1997multitask}
R.~Caruana, ``Multitask learning,'' {\em Machine learning}, vol.~28, no.~1, pp.~41--75, 1997.

\bibitem{ruder2017overview}
S.~Ruder, ``An overview of multi-task learning in deep neural networks,'' {\em arXiv preprint arXiv:1706.05098}, 2017.

\bibitem{zhang2021survey}
Y.~Zhang and Q.~Yang, ``A survey on multi-task learning,'' {\em IEEE Trans. Knowledge and Data Eng.}, 2021.

\bibitem{cervino2021multiTSP}
J.~Cervino, J.~A. Bazerque, M.~Calvo-Fullana, and A.~Ribeiro, ``Multi-task reinforcement learning in reproducing kernel hilbert spaces via cross-learning,'' {\em IEEE Trans. Signal Proc.}, vol.~69, pp.~5947--5962, 2021.

\bibitem{cervino2021multi}
J.~Cervino, J.~A. Bazerque, M.~Calvo-Fullana, and A.~Ribeiro, ``Multi-task supervised learning via cross-learning,'' in {\em European Signal Process. Conf. (EUSIPCO)}, pp.~1381--1385, IEEE, 2021.

\bibitem{kato2008multi}
T.~Kato, H.~Kashima, M.~Sugiyama, and K.~Asai, ``Multi-task learning via conic programming,'' in {\em Advances in Neural Info. Process. Syst.}, pp.~737--744, 2008.

\bibitem{kato2009conic}
T.~Kato, H.~Kashima, M.~Sugiyama, and K.~Asai, ``Conic programming for multitask learning,'' {\em IEEE Trans. Knowledge and Data Eng.}, vol.~22, no.~7, pp.~957--968, 2009.

\bibitem{koppel2017proximity}
A.~Koppel, B.~M. Sadler, and A.~Ribeiro, ``Proximity without consensus in online multiagent optimization,'' {\em IEEE Trans. Signal Process.}, vol.~65, no.~12, pp.~3062--3077, 2017.

\bibitem{koppel2019parsimonious}
A.~Koppel, G.~Warnell, E.~Stump, and A.~Ribeiro, ``Parsimonious online learning with kernels via sparse projections in function space,'' {\em The Journal of Machine Learning Research}, vol.~20, no.~1, pp.~83--126, 2019.

\bibitem{sharma2017learning}
S.~Sharma, A.~Jha, P.~Hegde, and B.~Ravindran, ``Learning to multi-task by active sampling,'' {\em arXiv preprint arXiv:1702.06053}, 2017.

\bibitem{sener2018multi}
O.~Sener and V.~Koltun, ``Multi-task learning as multi-objective optimization,'' in {\em Int. Conf. Machine Learning}, pp.~525--536, 2018.

\bibitem{Evengiou2004regularized}
T.~Evgeniou and M.~Pontil, ``Regularized multi--task learning,'' in {\em Proceedings of the 10th ACM SIGKDD international conference on knowledge discovery and data mining}, pp.~109--117, 2004.

\bibitem{zhang2010convex}
Y.~Zhang and D.~Y. Yeung, ``A convex formulation for learning task relationships in multi-task learning,'' in {\em Conf. on Uncertainty in Artificial Intell., UAI 2010}, p.~733, 2010.

\bibitem{torrey2010transfer}
L.~Torrey and J.~Shavlik, ``Transfer learning,'' in {\em Handbook of research on machine learning applications and trends: algorithms, methods, and techniques}, pp.~242--264, IGI global, 2010.

\bibitem{rozada2023matrix}
S.~Rozada and A.~G. Marques, ``Matrix low-rank approximation for policy gradient methods,'' in {\em IEEE Intl. Conf. Acoust., Speech and Signal Process. (ICASSP)}, pp.~1--5, IEEE, 2023.

\bibitem{rozada2024tensor}
S.~Rozada, S.~Paternain, and A.~G. Marques, ``Tensor and matrix low-rank value-function approximation in reinforcement learning,'' {\em IEEE Trans. Signal Process.}, 2024.

\bibitem{rozada2024tensorb}
S.~Rozada and A.~G. Marques, ``Tensor low-rank approximation of finite-horizon value functions,'' in {\em IEEE Intl. Conf. Acoust., Speech and Signal Process. (ICASSP)}, pp.~5975--5979, IEEE, 2024.

\bibitem{watkins1992q}
C.~J. Watkins and P.~Dayan, ``Q-learning,'' {\em Machine learning}, vol.~8, no.~3-4, pp.~279--292, 1992.

\bibitem{bradtke1996linear}
S.~J. Bradtke and A.~G. Barto, ``Linear least-squares algorithms for temporal difference learning,'' {\em Machine learning}, vol.~22, no.~1-3, pp.~33--57, 1996.

\bibitem{mnih2015human}
V.~Mnih, K.~Kavukcuoglu, D.~Silver, A.~A. Rusu, J.~Veness, M.~G. Bellemare, A.~Graves, M.~Riedmiller, A.~K. Fidjeland, G.~Ostrovski, {\em et~al.}, ``Human-level control through deep reinforcement learning,'' {\em Nature}, vol.~518, no.~7540, pp.~529--533, 2015.

\bibitem{schulman2017proximal}
J.~Schulman, F.~Wolski, P.~Dhariwal, A.~Radford, and O.~Klimov, ``Proximal policy optimization algorithms,'' {\em arXiv:1707.06347}, 2017.

\bibitem{mnih2013playing}
V.~Mnih, K.~Kavukcuoglu, D.~Silver, A.~Graves, I.~Antonoglou, D.~Wierstra, and M.~Riedmiller, ``Playing {A}tari with deep reinforcement learning,'' {\em arXiv preprint arXiv:1312.5602}, 2013.

\bibitem{sidiropoulos2017tensor}
N.~D. Sidiropoulos, L.~De~Lathauwer, X.~Fu, K.~Huang, E.~E. Papalexakis, and C.~Faloutsos, ``Tensor decomposition for signal processing and machine learning,'' {\em IEEE Trans. on Signal Processing}, vol.~65, no.~13, pp.~3551--3582, 2017.

\bibitem{geist2013algorithmic}
M.~Geist and O.~Pietquin, ``Algorithmic survey of parametric value function approximation,'' {\em IEEE Trans. Neural Netw.}, vol.~24, no.~6, pp.~845--867, 2013.

\bibitem{brockman2016openai}
G.~Brockman, V.~Cheung, L.~Pettersson, J.~Schneider, J.~Schulman, J.~Tang, and W.~Zaremba, ``Open{AI} {G}ym,'' {\em arXiv preprint arXiv:1606.01540}, 2016.

\bibitem{rozada2024code}
S.~Rozada, ``A tensor low-rank approximation for value functions in multi-task reinforcement learning.'' \url{https://github.com/sergiorozada12/multi-task-lrrl}, 2024.

\end{thebibliography}
%%%%%%%%%%%%%%%%%%%%%%%%%%%%%%%%%%%%%%%%%%

\end{document}